\documentclass{iopjournal}
\usepackage{adjustbox}
\usepackage{longtable}
\usepackage{url}
\usepackage{needspace}
\usepackage{natbib}
\usepackage{listings}

\begin{document}

\articletype{Article type} %	 e.g. Paper, Letter, Topical Review...

\title{Scale-Aware Adversarial Analysis: A Diagnostic for Generative AI in Multiscale Complex Systems}

\author{Mengke Zhao$^{1,3}$\orcid{0000-0003-0596-6608}, Guang-Xing Li$^{2,*}$\orcid{0000-0003-3144-1952}, Duo Xu$^{4,5}$\orcid{0000-0001-6216-8931} and Keping Qiu$^{1,3}$\orcid{0000-0002-5093-5088}}

% \author{Anonymous Authors}

\affil{$^1$School of Astronomy and Space Science, Nanjing University, 163 Xianlin Avenue, Nanjing 210023, Jiangsu, People’s Republic of China}

\affil{$^2$South-Western Institute for Astronomy Research, Yunnan University, Kunming 650091, People’s Republic of China}
\affil{$^3$Key Laboratory of Modern Astronomy and Astrophysics (Nanjing University), Ministry of Education, Nanjing 210023, Jiangsu, People’s Republic of China}

\affil{$^4$Canadian Institute for Theoretical Astrophysics, University of Toronto, 60 St. George Street, Toronto, ON M5S 3H8, Canada}

% % \affil{$^*$gxli@ynu.edu.cn,ligx.ngc7293@gmail.com}

\email{$^*$gxli@ynu.edu.cn,ligx.ngc7293@gmail.com}


\begin{abstract}
Complex physical systems, from supersonic turbulence to the macroscopic structure of the universe, are governed by continuous multiscale dynamics. 
While modern machine learning architectures excel at mapping the high-dimensional observables of these systems, it remains unclear whether they internalize the governing physical laws or merely interpolate discrete statistical correlations. 
Standard Explainable AI (XAI) architectures, particularly perturbation-based and gradient-saliency methods, rely on pixel-wise perturbations, which generate unphysical artifacts and push inputs off the valid empirical distribution.
To resolve this, we introduce a diagnostic framework driven by Constrained Diffusion Decomposition (CDD), a diffusion-based multiscale data decomposition algorithm that enables physically constrained data generation and model evaluation via scale-aware modifications. 
Applying this framework to a Denoising Diffusion Probabilistic Model (DDPM), we execute deterministic interventions directly within the continuous, CDD-based scale space.  
We demonstrate that under moderate physical perturbations, the unconstrained generative model exhibits localized structural freezing and non-linear instability rather than continuous PDE-like responses. 
The network fails to maintain cross-scale continuity, causing the generative trajectory to diverge when pushed into unseen physical states. 
By synthesizing a continuum of physically coherent states, this scale-informed methodology establishes a controlled test ground to evaluate algorithmic vulnerabilities, providing the rigorous physical constraints necessary for future architectures to respect the multiscale causality of the natural universe. 
\end{abstract}

\section{INTRODUCTION}

Complex physical systems in nature, from supersonic interstellar gas and chaotic turbulence \citep{2004RvMP...76..125M} to the macroscopic structure of the universe, are governed by continuous multiscale fluid dynamics \citep{1941DoSSR..32...16K,1981MNRAS.194..809L}. 
Consequently, these strongly non-linear environments serve as rigorous testbeds for computational physics \citep{1987ARA&A..25...23S,2007ARA&A..45..565M,2019MNRAS.490.3061V}. 
Any theoretical formulation or numerical solver attempting to capture these systems must preserve the cross-scale causality inherent to this continuous hierarchy. 
While deep generative models are increasingly deployed to map these high-dimensional observables, a fundamental vulnerability persists: these data-driven architectures function primarily as advanced morphological interpolators. 
Their capacity to genuinely internalize the multiscale causal mechanics of complex systems remains fundamentally deficient. 
It is strictly unknown whether modern architectures mathematically internalize the continuous Partial Differential Equations (PDEs) governing the underlying fluid dynamics, or merely interpolate discrete statistical textures.

Recently, deep learning and data-driven architectures have been increasingly deployed to map the high-dimensional observables of complex non-linear fluids into physical structures \citep{2020AnRFM..52..477B}.
However, a fundamental limitation remains: while deep learning can interpolate within well-posed observational distributions, it is unknown whether these models mathematically internalize the continuous Partial Differential Equations (PDEs) governing the underlying fluid dynamics, or merely memorize discrete statistical textures \citep{2018arXiv180608734R,2021IEEEP.109..704B}. 
Resolving this demands a transition from statistical fidelity metrics to diagnoses of physical causality. 
Although foundational Explainable AI (XAI) protocols establish mathematically rigorous attribution frameworks, their geometric application, particularly within perturbation-based methods and gradient-saliency algorithms, fundamentally relies on discrete, scale-agnostic pixel masking to compute spatial saliency \citep{2016arXiv160204938T,2017arXiv170507874L}.
In fluid dynamics, however, such geometrically unconstrained noise breaks the scale-space continuity dictated by the underlying PDEs \citep{2021NatRP...3..422K}. 
Arbitrary pixel-wise perturbations inject artificial singularities that violate continuous hydrodynamic gradients. 
Forcing the input into such unphysical regimes merely probes a network's blind statistical extrapolation, rendering these methods invalid for auditing true physical causality \citep{2021IEEEP.109..612S}. 

Capturing the structural hierarchy of complex fluids requires an appropriate representation basis. 
While traditional Fourier transforms assume global linear wave superposition, and standard continuous wavelets rely on zero-mean oscillatory basis functions, both introduce severe spectral artifacts when applied to the highly non-linear, sharp density contrasts characteristic of supersonic turbulence \citep{1999wtsp.book.....M}.
Applying such oscillatory or band-limited filters to these localized structures triggers the Gibbs phenomenon, introducing unphysical ringing and negative densities that violate fundamental conservation laws \citep{2002fvmh.book.....L,2009rsnm.book.....T}.
Therefore, executing structurally valid interventions necessitates a specialized physical projection operator. 
To resolve this, we introduce a scale-informed diagnostic framework utilizing Constrained Diffusion Decomposition \citep[CDD;][]{2022ApJS..259...59L}. 
Governed by continuous physical diffusion equations, CDD is a deterministic multiscale data decomposition algorithm that isolates spatial structures while strictly guaranteeing exact mass conservation and geometric non-negativity. 
Operating within this CDD-based physical scale space allows us to execute precise, geometry-constrained modifications, establishing a mathematically sound environment for deterministic data generation and causal model evaluation.

In this work, we introduce a scale-informed diagnostic framework that performs physical interventions through the continuous CDD scale space. 
By injecting scale-specific perturbations and evaluating the spatial structural response, we establish a transparent mechanism to audit the physical causality of data-driven networks. 
We deploy this framework to evaluate a representative Denoising Diffusion Probabilistic Model (DDPM; \citealt{2023ApJ...950..146X}).
To establish physical boundary conditions without relying on idealized synthetic data, we utilize an observationally-derived, highly non-linear fluid manifold (the NGC 1333 region; \citep{2008hsf1.book..308B,2022MNRAS.514L..16L}) as our empirical testbed. 
By diagnosing the scale-space continuity of this baseline architecture, we expose the structural limitations of unconstrained neural networks, ultimately establishing CDD-based scale-space continuity as the mathematical signature for physically consistent deep learning.

\begin{figure}
    \centering
    \includegraphics[width=\linewidth]{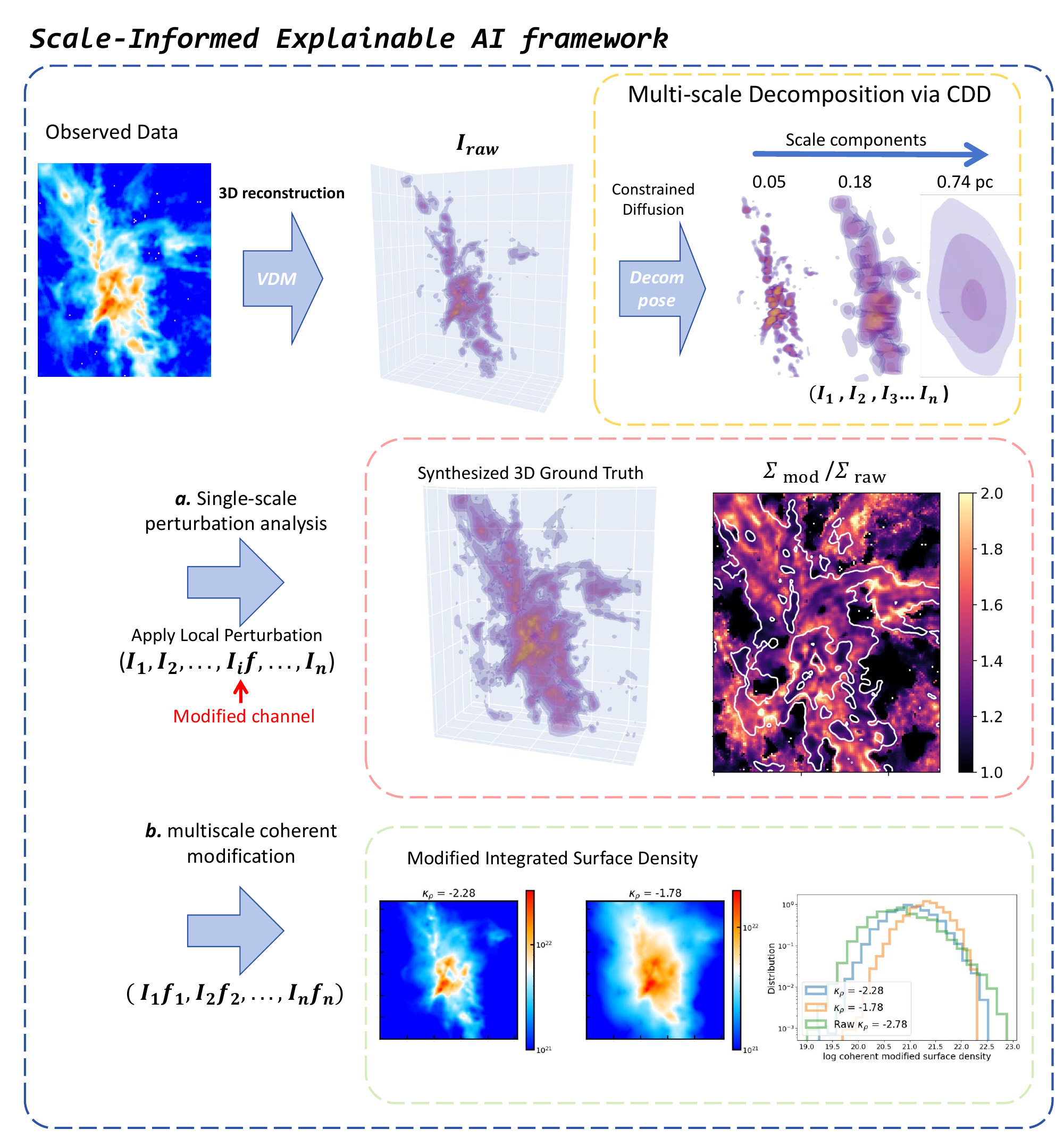}
    \caption{{\bf Construction of the Scale-Informed Diagnostic Framework.}
    Our experimental pipeline begins by reconstructing a 3D latent density volume from a background-subtracted observational integrated surface density map ($\Sigma_{\rm obs}$) using the Volume Density Mapper (VDM), establishing a mass-conserving deterministic physical baseline. 
    This continuous volume is subsequently projected into discrete spatial frequencies via Constrained Diffusion Decomposition (CDD). 
    From this rigorous scale-space representation, we synthesize 3D ground truths ($I_{\rm mod}$) through two distinct intervention mechanisms: (1) \textit{Single-Scale Perturbation}, which injects deterministic magnifications into isolated spatial components, and (2) \textit{Multiscale Coherent Modification}, which systematically rotates the underlying scaling cascade to simulate physical phase transitions (e.g., shifting the density profile). 
    These modified 3D volumes are then projected to generate the perturbed 2D observables ($\Sigma_{\rm mod}$). 
    This closed-loop geometric system provides a transparent causal pair ($\Sigma_{\rm mod} \rightarrow I_{\rm mod}$) to rigorously probe the scale-space structural continuity of deep generative architectures.}
    \label{fig1}
\end{figure}

\section{THE SCALE-INFORMED FRAMEWORK: CAUSAL INTERVENTION AND STATE SPACE EXPANSION}

In multiscale complex physical systems, continuous hierarchical structures are the direct manifestations of the underlying non-linear partial differential equations (PDEs). 
Standard computer vision perturbations (e.g., arbitrary pixel-wise noise) structurally violate these governing PDEs, pushing inputs into unphysical states \citep{2021NatRP...3..422K}.
To overcome the limitations of purely data-driven black-box models, we introduce a scale-informed Explainable AI (XAI) framework. 
We employ Constrained Diffusion Decomposition \citep[CDD;][]{2022ApJS..259...59L}, a deterministic $N$-dimensional operator that decomposes arbitrary physical fields into a complete hierarchy of characteristic spatial scales. 
This operation strictly enforces geometric non-negativity and absolute mass conservation, mathematically guaranteeing that the superposition of all isolated exactly converges to the original input.
This mathematically guarantees that all structural interventions remain physically legal.

\subsection{Establishing the Deterministic Baseline and Scale Decomposition}

The diagnostic procedure begins with a 2D integrated observable field (e.g., a continuous surface density map). 
Utilizing the Volume Density Mapper \citep[VDM;][]{2025arXiv250917369L,2026ApJ...997..345Z}, a deterministic inverse solver, we reconstruct a physically-consistent 3D volume density representation, $I_{\rm raw}(\mathbf{r})$. 
VDM enforces mass conservation and physical power spectra of the observation, providing a mathematically coherent 3D physical baseline.

We then apply the CDD operator directly to this 3D structure, projecting it into a discrete sequence of 3D spatial scale sub-spaces:
\begin{equation}
 I_{\rm raw}(\mathbf{r}) = \sum_{i=1}^{N} I_i(\mathbf{r}; r_i)
\end{equation}
where $I_i(\mathbf{r}; r_i)$ represents the isolated 3D density field containing structures exclusively at the characteristic spatial scale $r_i$.
From this scale-space representation, our framework executes two mathematically defined operations: (1) Single-Scale Perturbation for AI diagnostics, and (2) Multiscale Coherent Modification for state space expansion.

\subsection{Single-Scale Perturbative Modifications}

To probe whether a generative AI model has internalized physical causality, we directly evaluate its structural response to moderate interventions in continuous scale space.
We select a specific target scale $r_j$ and inject a deterministic magnification factor $f$ into its corresponding density component, leaving the orthogonal scales strictly invariant:
\begin{equation}\label{eq2}
I_{\rm mod}(\mathbf{r}) = \sum_{i \neq j} I_i(\mathbf{r}; r_i) + f \cdot I_j(\mathbf{r}; r_j) = I_{\rm raw}(\mathbf{r}) + (f-1) \cdot I_j(\mathbf{r}; r_j)
\end{equation}
This perturbed 3D volume, $I_{\rm mod}(\mathbf{r})$, acts as our controlled physical target.
The factor $f$ serves as a controlled causal intervention (e.g., $f \in \{1.01, 1.1, 1.5, 3.0\}$). We apply a spatial projection operator to integrate this volume, yielding the modified 2D observable: $\Sigma_{\rm mod}(x, y) = \int I_{\rm mod}(\mathbf{r}) dz$.

By comparatively feeding both the unperturbed ($\Sigma_{\rm raw}$) and perturbed ($\Sigma_{\rm mod}$) fields into the pre-trained generative architecture, we compute the relative structural response ratio ($I_{\rm pred, mod} / I_{\rm pred, raw}$). 
This empirical ratio acts as a deterministic discrete proxy for the network's local scale-space sensitivity, providing a rigorous metric to evaluate structural continuity \citep{2018arXiv180208760N, 2021IEEEP.109..704B}. 
This explicit mapping allows us to diagnose whether the network maintains continuous structural monotonicity or suffers from non-linear physical instability when confronting specific spatial frequencies.

\subsection{Multiscale Coherent Modification}\label{sec2.3}

The structural evolution of a complex fluid system is encoded in the multiscale scaling laws of its density field \citep{1981MNRAS.194..809L,2012ApJ...761..156F}. 
While numerical simulations offer high-fidelity physical insights at extreme computational costs \citep{2021MNRAS.506.2199G}, and standard data augmentation relies on statistical rather than physical priors \citep{2019JournalBigData_Shorten}, our framework provides an efficient geometric alternative. By mathematically tilting the governing physical cascade directly within the discrete scale space, we expand the valid empirical data manifold.

We apply a coherent scale-dependent operator, parameterized by an index $s_c$, across all spatial components simultaneously:
\begin{equation}\label{eq3}
I_{\rm aug}(\mathbf{r}) = \sum_{i=1}^{N} I_i(\mathbf{r}; r_i) \cdot \left( \frac{r_i}{r_{\rm ref}} \right)^{s_c}
\end{equation}
where $r_{\rm ref}$ acts as a macroscopic boundary condition to prevent unphysical mass divergence. 
This operation shifts the continuous density exponent from its raw physical state ($\kappa_\rho$, \citealt{2022MNRAS.514L..16L}) to an evolved state ($\kappa_\rho + s_c$). 

Summing these re-weighted components yields a modified 3D volume, $I_{\rm aug}(\mathbf{r})$, and its 2D projection, $\Sigma_{\rm aug}(x, y)$. 
This scale-space rotation directly shifts the density cascade exponent ($\kappa_\rho$), mathematically emulating the continuous structural transition between steep, gravity-driven non-linear aggregations and diffuse, turbulence-dominated scale-invariant fractals (Appendix\,\ref{ApB}). 
Because high-fidelity empirical measurements of complex fluid systems are inherently sparse, this geometric rotation serves as a deterministic, physically constrained interpolator.
It synthesizes a continuum of valid structural states, expanding the empirical data manifold under strict physical constraints.

\begin{figure}
    \centering
    \includegraphics[width=\linewidth]{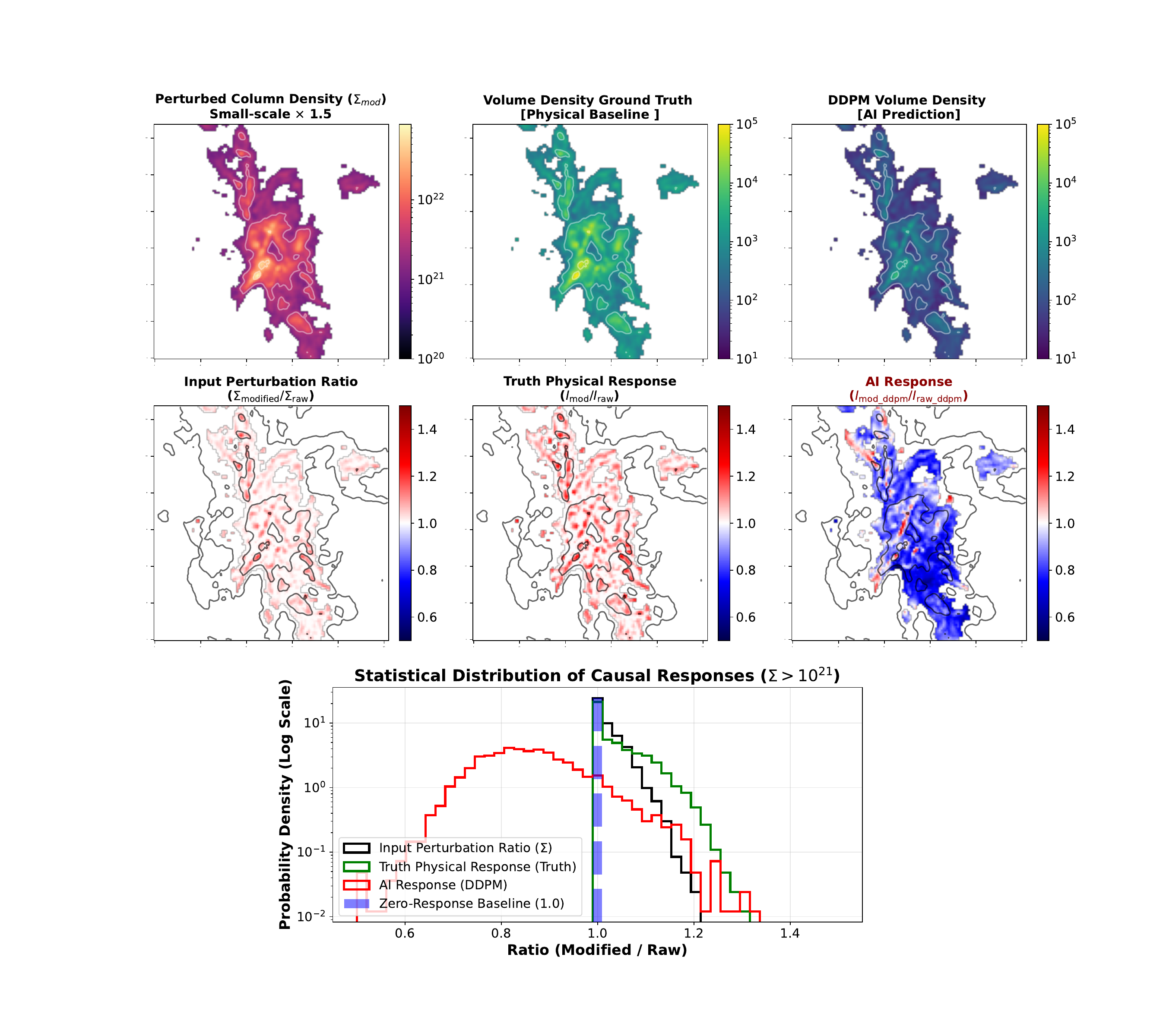}
    \caption{\textbf{The Negative Response Paradox and Violation of structural Monotonicity.} 
    Spatial and statistical distributions of the volume density response ratio ($I_{\rm mod}/I_{\rm raw}$) following a deterministic single-scale mass intervention ($f = 1.5$) applied to the physical baseline. 
    \textbf{(a) Deterministic Physical Baseline:} The causal response yields a positive density enhancement ($> 1.0$, red regions). 
    This confirms the strict structural monotonicity inherent to the physical manifold. 
    \textbf{(b) DDPM Prediction:} The deep generative model predicts an unphysical density depletion ($< 1.0$, blue regions) across the intervened area. The accompanying histograms quantitatively capture this anti-physical shift across the zero-response baseline ($1.0$).}
    \label{fig2}
\end{figure}

\begin{figure}
    \centering
    \includegraphics[width=\linewidth]{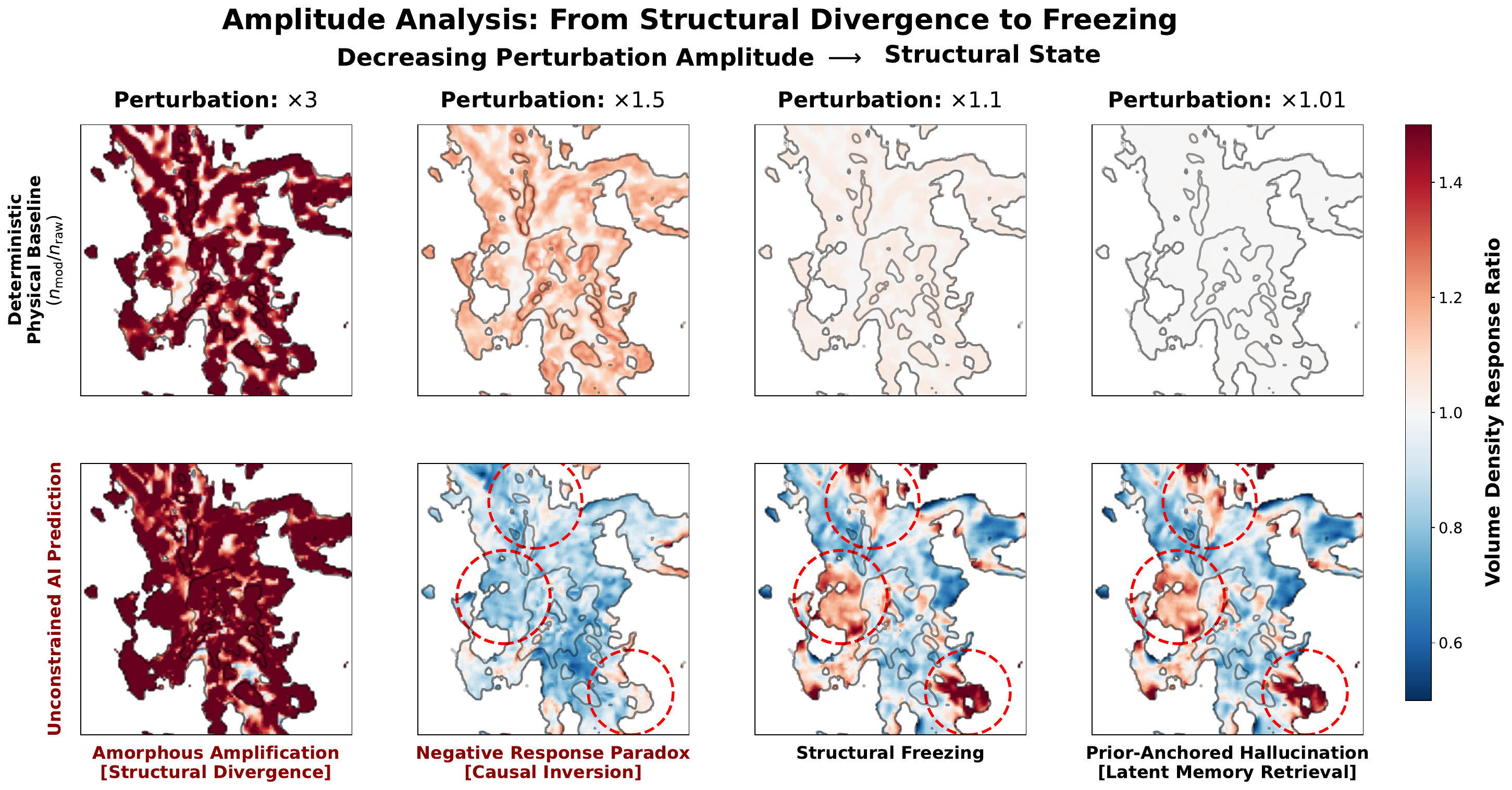}
    \caption{\textbf{Amplitude Diagnostics: Topological Freezing and Prior-Anchored Hallucinations.} 
    Evaluation of the diffusion manifold's topological resilience by scaling the mass perturbation magnitude from macroscopic ($f=3.0$) to microscopic ($f=1.01$) limits. 
    \textbf{Top row (Deterministic Physical Baseline):} The physical operator maintains continuous sensitivity, scaling the density enhancement while preserving the causal topological skeleton. 
    \textbf{Bottom row (DDPM Prediction):} 
    The generative model exhibits severe non-linear phase transitions. 
    Under macroscopic intervention ($f=3.0$), the model exhibits structural divergence. 
    Crucially, as the perturbation magnitude decays below a critical threshold ($f \le 1.1$), the model enters a structurally frozen state. 
    Instead of converging to equilibrium, the local structural response stalls, and the network defaults to its unconditioned prior. 
    The invariant density anomaly (dashed circles) is diagnosed as a prior-anchored hallucination rather than a continuous physical derivative.}
    \label{fig4}
\end{figure}

\section{DIAGNOSING CAUSALITY VIA SCALE-SPACE INTERVENTIONS}

To evaluate our scale-informed framework, we audit a representative Denoising Diffusion Probabilistic Model (DDPM; \citealt{2023ApJ...950..146X}). 
This architecture addresses the ill-posed inverse problem \citep{2005SIAM_Tarantola} of mapping 2D observables to 3D density fields. 
Deep neural networks demonstrate high empirical fidelity in fluid reconstruction tasks \citep{Holl2024TheUE}, yet their structural stability under scale-space interventions remains largely untested. 

We select a U-Net diffusion architecture \citep{Ho2020DenoisingDP} to establish the structural baseline. 
U-Net's hierarchical convolutions naturally approximate local finite-difference stencils \citep{2017arXiv171009668L}, providing an inductive bias for spatial continuity. 
Conversely, architectures relying on patch tokenization \citep{2022arXiv220206709P,2022arXiv220400993B} act as low-pass filters that artificially smooth high-frequency boundaries, such as shockwaves.  
Diagnosing this constrained U-Net provides a controlled environment to test physical causality.

For the physical boundary condition, we utilize a continuous fluid manifold governed primarily by self-gravity and supersonic turbulence \citep{2022MNRAS.514L..16L}. 
By explicitly avoiding domains disrupted by discontinuous, macroscopic energy injections (e.g., massive stellar feedback), we isolate the core multiscale cascade \citep{2008hsf1.book..308B}. 
This ensures the generative solver is tested strictly on physically possible data configurations, where structural validity is explicitly measured by their continuity within the CDD-based scale space. 
We establish the deterministic 3D physical baseline by reconstructing the volume density field via the Volume Density Mapper (VDM; \citealt{2025arXiv250917369L}). 

\subsection{Single-Scale Perturbation: The Negative Response Paradox}\label{sec3.1}

A physical generative solver must preserve structural monotonicity: a localized mass enhancement in the input must yield a positive density increment in the output. 
We intervene on the physical baseline by injecting a deterministic magnification ($f=1.5$) at isolated spatial frequencies. 
Because turbulence is scale-free \citep{2004ARA&A..42..211E}, this moderate structural perturbation mimics a natural cascade variation that falls within the empirical uncertainty margins of complex fluid measurements, ensuring the intervention is physically permissible.

The diagnostic reveals a "negative response paradox" (see Figure\,\ref{fig2}). 
While the physical baseline requires a mass increment, the DDPM predicts a mass depletion within the intervened regions. 
The model computes a negative spatial derivative, generating an anti-physical density void. 
From the perspective of Causal Representation Learning \citep{2021IEEEP.109..612S}, this confirms that the architecture purely interpolates the observational joint probability distribution without internalizing the causality of mass assembly. 
As detailed in Section\,\ref{sec3.2}, this structural discontinuity stems from the fragmentation of the learned physical representation \citep{2021arXiv211009485B}.

\begin{figure}[h]
    \centering
    \includegraphics[width=\linewidth]{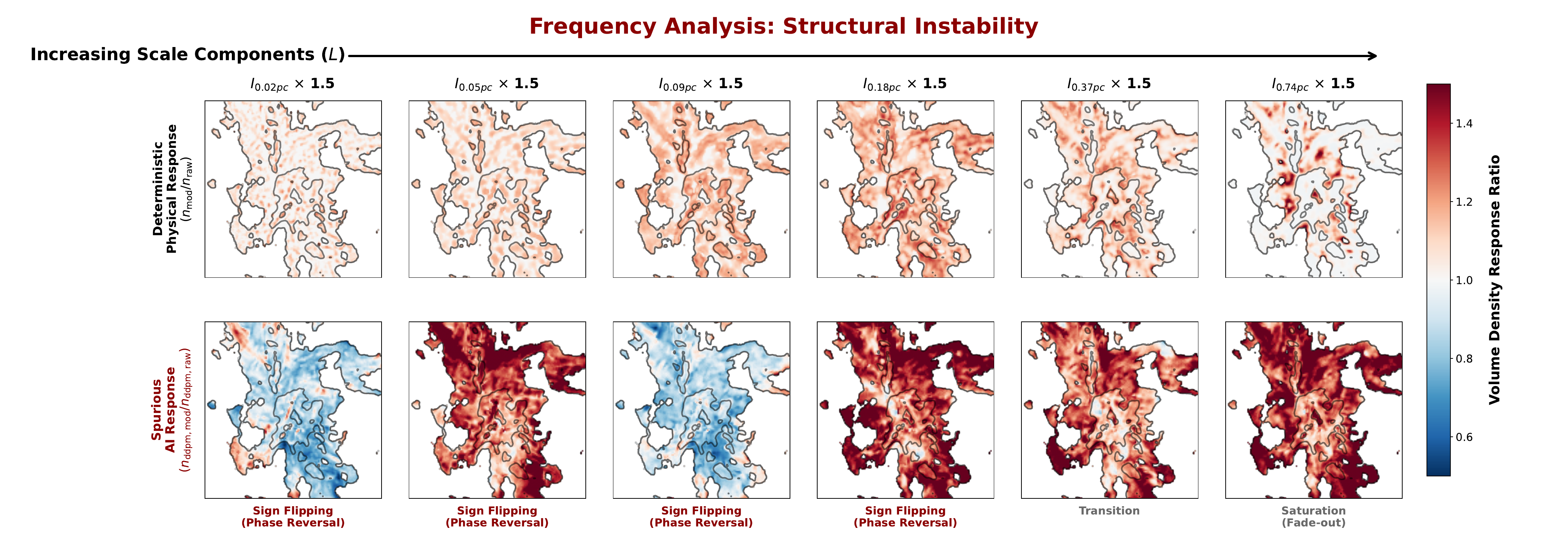} 
    \caption{\textbf{Geometric Diagnosis of Score Field Divergence and Manifold Fragmentation.} 
    Controlled single-scale mass perturbations ($f = 1.5$) are injected independently across isolated spatial frequencies using CDD. 
    \textbf{Top panels (Deterministic Physical Baseline):} The baseline exhibits a universally positive density amplification across the continuous scale space. 
    \textbf{Bottom panels (DDPM Prediction):} The generative model exhibits structural instability. 
    As the intervention scans across spatial frequencies, the local structural derivative flips polarity. 
    This alternating structural response, oscillating between positive mass increments and anti-physical mass depletions, geometrically confirms that the conditional trajectory is puncturing through spurious non-physical boundaries where the learned structural gradients severely diverge within a fragmented learned physical representation.}
    \label{figCrazy}
\end{figure}

\subsection{Frequency and Amplitude Analysis: Diagnosing Structural Freezing and Scale-Space Instability}\label{sec3.2}

To demonstrate the diagnostic capability of our scale-informed framework, we first evaluate the network's response to diminishing intervention amplitudes. 
Physically, a continuous fluid system requires the structural response to smoothly converge to zero as the perturbation vanishes. 
However, our diagnostic pipeline reveals severe non-linear failure modes in the baseline DDPM across the amplitude spectrum (Figure\,\ref{fig4}). 
At large perturbations ($f=3.0$), the network fails to confine the structural response, generating unstructured density artifacts that bleed beyond the spatial footprint of the intervention.
Conversely, in the micro-perturbation limit ($f \le 1.1$), the structural response completely freezes. 
A localized density anomaly (dashed circles in Figure\,\ref{fig4}) becomes entirely insensitive to the vanishing physical perturbation. 
Instead of maintaining a continuous structural derivative, the network diverges to a pre-learned discrete state, exposing severe statistical memorization.
By quantifying this response, the framework exposes that without the continuous regularization of governing PDEs, the network suffers from severe structural freezing. 
Forced into blind statistical extrapolation, the model decouples from the physical input and defaults to unphysical structural hallucinations \citep{adcock2019instabilitiesdeeplearning, 2021RSPTA.37900093K}.

Beyond amplitude, the framework's unique capacity to isolate continuous spatial domains allows us to evaluate the structural sensitivity across different frequencies. 
A physically consistent fluid emulator must maintain a strictly positive structural derivative across the entire cascade. 
We inject a constant perturbation ($f=1.5$) independently across discrete spatial frequencies, spanning the non-linear cascade from macroscopic integral scales down to the localized dissipation limit (For strict reproducibility, the physical boundaries of this specific manifold span $L \in [0.05, 0.74]$ pc).
This frequency-scanning diagnostic captures the unphysical consequences of unconstrained generative mapping (Figure\,\ref{figCrazy}). 
The physical baseline (top panels) exhibits a smooth, non-oscillatory density amplification, reflecting the continuous fluid cascade. 
In contrast, the DDPM (bottom panels) exhibits an erratic oscillatory response. 
As the intervention frequency shifts, the network's structural derivative abruptly inverts its sign, alternating between positive mass increments and unphysical mass depletions across adjacent scale layers.

Through the lens of our scale-informed diagnostic, we empirically demonstrate that unconstrained networks fail to maintain cross-scale continuity, the fundamental signature of complex physical systems.
Injecting perturbations across isolated spatial frequencies pushes the input into unseen physical states, causing the generative trajectory to diverge and abruptly invert the local structural gradient \citep{2021arXiv211009485B, 2016arXiv160605336R}. 
By quantifying these physical discontinuities, our framework demonstrates that mitigating such algorithmic instability requires enforcing strict scale-space continuity during the data generation process (Section\,\ref{sec2.3}).

\section{CONCLUSION: SCALE-SPACE CONTINUITY AS A FUNDAMENTAL ALIGNMENT METRIC}

The explicit analytical formulation of multiscale complex systems is computationally intractable. 
While deep generative architectures provide a scalable mechanism to encode these nonlinear physical structures, it remains unverified whether unconstrained, data-driven models internalize the underlying physical causality.
To bridge this gap, we introduced a Scale-Informed Explainable AI (XAI) framework driven by a multiscale data decomposition algorithm, Constrained Diffusion Decomposition \citep[CDD;][]{2022ApJS..259...59L}.

The defining characteristic of multiscale fluid dynamics is cross-scale causal continuity. 
Rather than relying on unclosed fluid PDEs, the scale-informed XAI framework provides a deterministic geometric mechanism that respects this scale-space continuity.
It generates physically meaningful structures via scale-informed modifications to the input data, establishing a paradigm where structural differences between synthesized states can be controlled by user-defined inputs, targeting spatial scale, perturbation amplitude, and scale-space distribution.

This scale-informed framework operates through two core application branches:
\begin{enumerate}
\item \textbf{Scale-Based Model Evaluation:} 
By injecting controlled, scale-specific perturbations, the framework generates differential data pairs whose intrinsic physical consistency is strictly guaranteed by CDD-based scale-space continuity. 
We establish that evaluating the network's structural response to these continuous, paired inputs provides a definitive causal metric to audit AI performance and isolate non-linear failure modes. 
\item \textbf{Scale-Informed Data Generation:} 
The framework explicitly synthesizes mathematically coherent physical states by re-weighting the natural multiscale cascade \citep[$\kappa_\rho$;][]{2018MNRAS.477.4951L,2022MNRAS.514L..16L}. 
This geometric operation expands the valid empirical manifold, enabling the generation of data with strictly controlled continuity and resolving observational sparsity. 
\end{enumerate}

Applying this scale-informed framework to a baseline diffusion architecture \citep{2023ApJ...950..146X}, we discovered the vulnerabilities of pure data-driven interpolation. 
The diagnostic exposed algorithmic failures: spatial divergence, localized structural freezing, and non-physical oscillatory instability. 
These failures confirm that when pushed into unseen physical states, unconstrained networks fail to resolve continuous hydrodynamic processes, defaulting instead to unphysical statistical hallucinations.

This work establishes a CDD-based scale-space view of generative model performance on complex physical structures. 
Deep learning models applied to complex multiscale systems must be mathematically constrained by cross-scale continuity. 
By enforcing scale-space continuity as a fundamental alignment metric, we provide the essential geometric priors required for future architectures to overcome structural freezing and capture the continuum mechanics of complex physical systems.

\bibliographystyle{aasjournalv7}
\bibliography{reference}

\newpage

\appendix
\section{Geometric Formulation of Continuous State Space Expansion}\label{ApB}

\begin{figure}[h]
    \centering
    \includegraphics[width=0.6\linewidth]{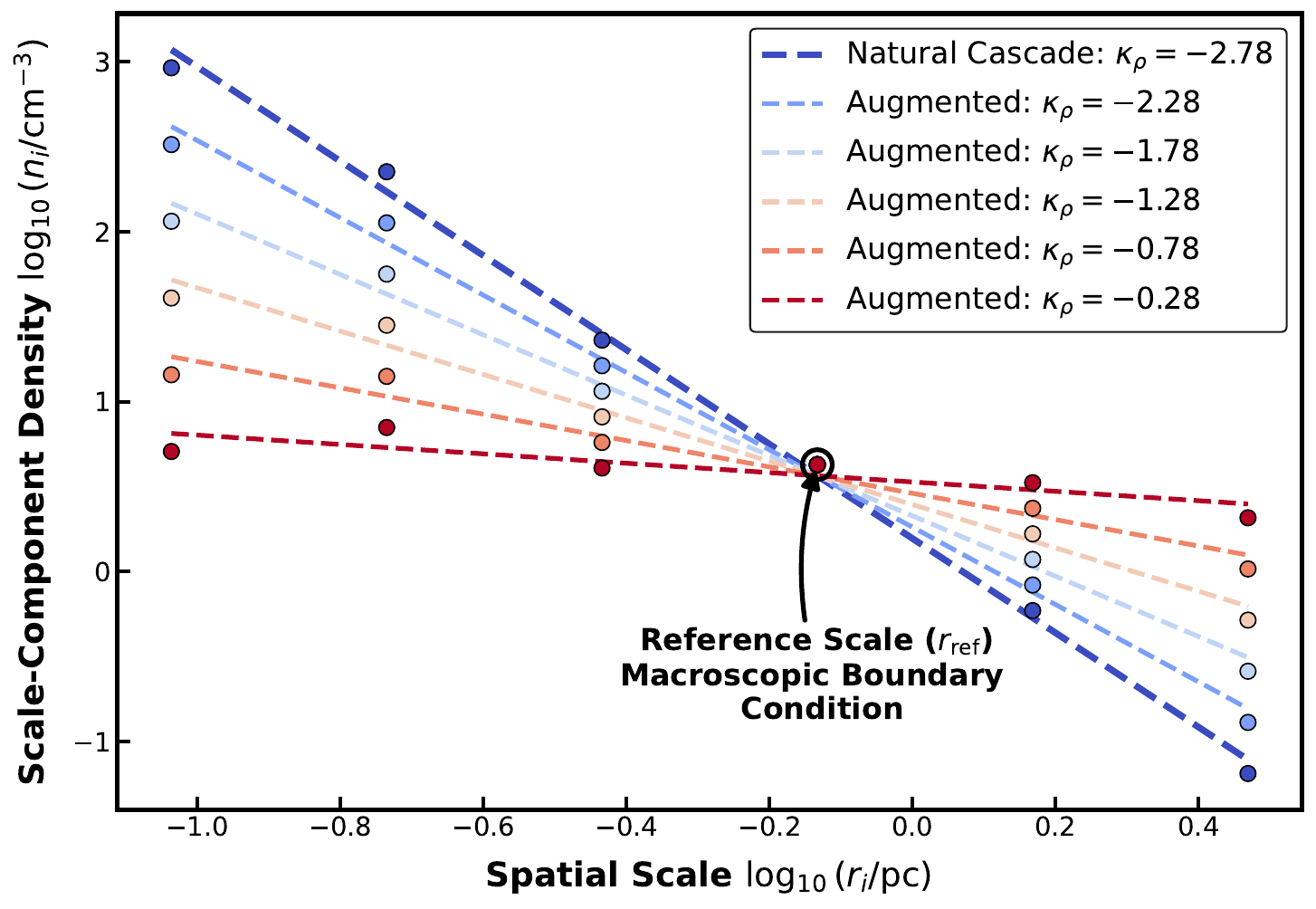}
    \includegraphics[width=0.9\linewidth]{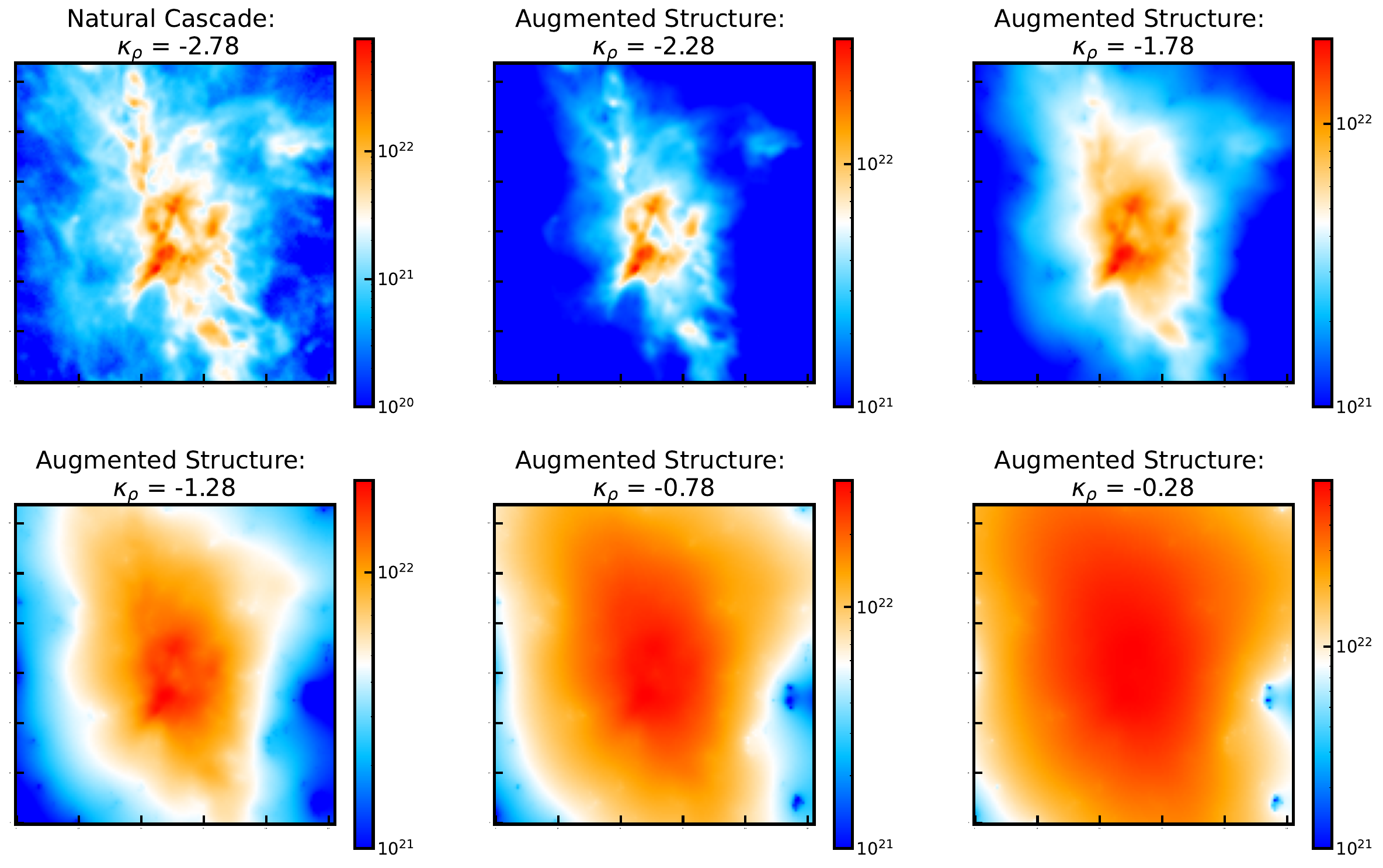}
    \caption{Mathematical formulation and morphological manifestation of multiscale coherent modification.
    \textbf{(a) Geometric Rotation in CDD-based Scale Space:} The scale-component density $n_i$ versus spatial scale $r_i$ is plotted in the density-scale phase space. The augmentation pipeline (Equation\,\ref{eq3}) rotates the natural scaling cascade ($\kappa_\rho = -2.78$) around a macroscopic boundary condition ($r_{\rm ref}$). Within the physically permissible dynamic range, the rotated cascade strictly maintains scale-space continuity.
    \textbf{(b) Morphological Phase Transitions:} 
    2D integrated surface density projections tracking the corresponding 3D modifications. The log-space rotation maps directly to a physical phase transition, mathematically simulating the continuous structural evolution between steep, gravity-driven non-linear aggregations and diffuse, turbulence-dominated fractals, while preserving physical causality.}
    \label{figB}
\end{figure}

To illustrate the multiscale coherent tilting introduced in Section\,\ref{sec2.3}, Figure\,\ref{figB} dissects the Constrained Diffusion Decomposition (CDD) pipeline from both scale-space geometry and physical morphology.

Figure\,\ref{figB}a illustrates the intervention in the density-scale phase space. 
The deterministic physical baseline of the highly non-linear fluid manifold exhibits a continuous cascade characterized by a fundamental density exponent of $\kappa_\rho = -2.78$. 
By applying the scale-dependent operator (Equation\,\ref{eq3}), our framework executes a linear transformation, realizing a coherent rotation of the scaling cascade.

This structural rotation in phase space is regulated by the reference scale $r_{\rm ref}$, which acts as a macroscopic boundary condition. 
Serving as a fixed physical pivot, this macroscopic boundary restricts the scaling transformation, guaranteeing that flattening the cascade does not induce unphysical mass divergence at microscopic scales. 
This geometric constraint ensures scale-space continuity across the newly synthesized physical states.

Figure\,\ref{figB}b demonstrates the spatial manifestation of this geometric operation. 
As the density exponent is flattened from the steep \citep{2022MNRAS.514L..16L}, gravity-dominated regime ($\kappa_\rho = -2.78$) to the shallow, turbulence-dominated hierarchy ($\kappa_\rho = -0.28$), the observable undergoes a continuous morphological phase transition. 
The localized, high-contrast density peaks are structurally redistributed into a diffuse, scale-invariant fractal network.
This morphological evolution preserves the strict physical causality of the scale space.

By mathematically rotating the multiscale geometry, this framework provides a deterministic, closed-form operator to synthesize structurally continuous states. 
Unlike standard numerical hydrodynamic simulations, where coupled non-linear PDE solvers make it mathematically impossible to isolate and smoothly perturb a single structural scaling parameter, this geometric framework executes an exact, orthogonal intervention on the density cascade. 
It synthesizes a continuous spectrum of mathematically valid states, establishing the precise geometric priors required to rigorously audit generative networks and expand the physical data manifold.

\end{document}